\documentclass{article}

\usepackage{PRIMEarxiv}

\usepackage[round]{natbib}

\usepackage{amsmath,amsfonts,bm}
\usepackage{hyperref}
\usepackage{url}
\usepackage{graphicx}
\usepackage{xcolor}
\usepackage{enumerate}
\usepackage{textcomp}
\usepackage{mathrsfs}
\usepackage{amsthm}
\usepackage{bbm}
\usepackage{epigraph}
\usepackage{etoolbox}
\usepackage{thmtools, thm-restate}
\usepackage{proof-at-the-end}
 
\newtheorem{theorem}{Theorem}
\newtheorem{lemma}[theorem]{Lemma}

\newtheorem{corollary}[theorem]{Corollary}

\usepackage{algorithm} 
\usepackage{algpseudocode}
\usepackage{float}

\def\diffentropy{\bf h}

\def\E{\mathbb{E}}

\def\H{\mathbb{H}}
\def\diffentropy{\mathbf{h}}
\def\I{\mathbb{I}}
\def\Pr{\mathbb{P}}
\def\R{\mathbb{R}}
\def\1{\mathbf{1}}

\newcommand{\Rc}{\mathcal{R}}

\newcommand{\Ac}{\mathcal{A}}

\newcommand{\Dc}{\mathcal{D}}

\newcommand{\Oc}{\mathcal{O}}

\usepackage{hyperref}
\DeclareMathOperator*{\argmax}{arg\,max}
\DeclareMathOperator*{\argmin}{arg\,min}

\usepackage[mathscr]{euscript}
\hypersetup{
    colorlinks=true,
    linkcolor=blue,
    filecolor=magenta,      
    urlcolor=purple,
}
\usepackage{xcolor}

\newcount\Comments
\newcommand{\kibitz}[2]{\ifnum\Comments=1{\textcolor{#1}{\textsf{\footnotesize #2}}}\fi}
\usepackage{color}
\definecolor{darkred}{rgb}{0.7,0,0}
\definecolor{darkgreen}{rgb}{0.0,0.5,0.0}
\definecolor{darkblue}{rgb}{0.0,0.0,0.5}
\definecolor{teal}{rgb}{0.0,0.5,0.5}

\title{Aligning AI Agents via\\ Information-Directed Sampling
}

\author{
  Hong Jun Jeon \\
  Computer Science \\
  Stanford University \\
  Stanford, CA\\
  \texttt{hjjeon@stanford.edu} \\
   \And
  Benjamin Van Roy \\
  Stanford University \\
  Stanford, CA\\
  \texttt{bvr@stanford.edu} \\
}

\begin{document}
\maketitle

\begin{abstract}
    The staggering feats of AI systems have brought to attention the topic of AI Alignment: \emph{aligning} a ``superintelligent'' AI agent's actions with humanity's interests.  Many existing frameworks/algorithms in alignment study the problem on a myopic horizon or study learning from human feedback in isolation, relying on the contrived assumption that the agent has already perfectly identified the environment.  As a starting point to address these limitations, we define a class of \emph{bandit alignment problems} as an extension of classic multi-armed bandit problems.  A bandit alignment problem involves an agent tasked with maximizing long-run expected reward by interacting with an environment \emph{and} a human, both involving details/preferences initially unknown to the agent.  The reward of actions in the environment depends on both observed outcomes and human preferences.  Furthermore, costs are associated with querying the human to learn preferences.  Therefore, an effective agent ought to intelligently trade-off exploration (of the environment \emph{and} human) and exploitation.  We study these trade-offs theoretically and empirically in a toy bandit alignment problem which resembles the beta-Bernoulli bandit.  We demonstrate while naive exploration algorithms which reflect current practices and even touted algorithms such as Thompson sampling both fail to provide acceptable solutions to this problem, information-directed sampling achieves favorable regret.
\end{abstract}


\section{Introduction}
The staggering progress of artificial intelligence across the past decade has far exceeded anyone's expectations.  With AI agents toppling grand challenges ranging from unfathomably complex decision problems such as Go \citep{silver2017mastering}, to interacting seamlessly with humans via language \citep{achiam2023gpt}, uncertainty about the future capabilities of theses systems looms ahead.  As these agents begin to surpass humans in both knowledge and capabilities, the question of alignment becomes crucial: how do we ensure that these superintelligent agents have goals which are aligned with those of humanity?  To mitigate the unintended (potentially catastrophic) consequences of deploying a misaligned superintelligence, much work is required on both the theoretical and algorithmic fronts.

Many global corporations have taken this maxim to heart and perform an explicit ``post-training'' phase to better ``align'' their models with the goals of the human end user prior to deployment \citep{ouyang2022training, achiam2023gpt, team2023gemini, dubey2024llama}.  A natural question is: what are the limitations of the current alignment protocols?  A key concern which we highlight in this work is that this paradigm follows an explicit ``explore then exploit'' pattern.  Information is first gathered without an explicit sensitivity to costs and then it is leveraged without further exploration at deployment.  When the algorithm is deployed in such a manner for a sufficiently long horizon, the concerns surrounding alignment amplify.  What to do in the almost sure scenario that fatal edge cases are omitted in the ``explore'' phase?  A natural answer is to repeat this pattern of ``explore then exploit'' in a periodic fashion ad nauseam.  However, when analyzing this procedure in even simple problem settings, we demonstrate that it is \emph{highly suboptimal}.  This proposition becomes all the more relevant as the scale at which theses systems operate continues to grow.  We believe that these protocols arise from a framing of alignment which does not correctly account for cost and reward across a long horizon.  In this work, we demonstrate that by correctly attributing costs to querying humans about their preferences and rewards to achieving aligned outcomes, an agent which aims to maximize reward (cost is negative reward) will exhibit behavior that we would desire from a superintelligent agent.

We are not the first to propose frameworks for alignment.  Notably, cooperative inverse reinforcement learning (CIRL) \cite{hadfield2016cooperative} defines a broad class of problems which attempt to capture the essence of alignment.  A CIRL problem is a 2 player markov game involving a human and a robot in which the ladder does not know the reward.  We levy two main criticisms of this work 1) the human (and robot) have full knowledge of the environment and 2) the POMDP formulation obfuscates the search for scalable algorithms.  In the reality, neither humanity nor the superintelligent agent will be able to fully identify the environment.  Part of the process of aligning superintelligent agents ought to involve these agents accruing knowledge about the external environment beyond what it is bestowed at initialization, and beyond what is known to humanity.  As a result, in our formulation, we view the environment and the human as separate systems, both of which the agent must \emph{simultaneously} explore while also accruing enough reward.

While CIRL problems fall into a class of POMDP problems, we argue that this framing often confines the practitioner to considering only algorithms which involve (approximately) solving POMDPs.  While even simple multi-armed bandit problems can be formulated as POMDPs, POMDP solvers are hardly a practitioner's algorithm of choice (due to scalability). 
 Algorithms such as Thompson sampling \citep{thompson1933likelihood}, upper confidence bounds \citep{lai1985asymptotically} and information-directed sampling \citep{russo2014learning} have been demonstrated to perform well in bandit and reinforcement learning environments despite the fact that they perform somewhat heuristic exploration schemes when compared to solving a POMDP with a complex belief state.  Moreover, advances in scaleable uncertainty modeling via methods such as epinet \citep{osband2023epistemic} enable efficient approximate variants of TS and IDS for complex problems involving neural networks.  The accelerated prototyping afforded by these scaleable algorithms ought to facilitate advances in AI alignment on both algorithmic and theoretical fronts.

In this paper, we define a class of problems for the purposes of studying alignment of AI agents.  This class of \emph{bandit alignment problems} builds off of the hidden utility bandit formulation of \cite{freedman2023active}.  We provide one simple problem instance which we refer to as the beta-Bernoulli bandit alignment problem and provide a thorough theoretical and empirical analysis.  As its name suggests, it reflects the classic beta-Bernoulli bandit from the multi-armed bandit literature.  Despite its simplicity, the environment exhibits sufficient complexity to exhibit fascinating tradeoffs between exploiting accumulated information, exploring the environment, and learning about the human's preferences.  We demonstrate that naive ``explore then exploit'' algorithms are unable to achieve adequate performance in this environment, exhibiting regret lower bounds which are linear in the size of the action set and the time horizon.  Furthermore even Thompson sampling fails to produce a result which is sublinear in the horizon.  However, we demonstrate theoretically that Information Directed Sampling \citep{russo2014learning} achieves regret which is sublinear in both the action set size and the horizon.  Our results demonstrate the importance of designing algorithms which explore in a reward-sensitive manner.  We believe this work provides many exciting directions of future work in AI alignment, both theoretical and algorithmic.

\section{Probabilistic Framework}
We define all random variables with respect to a common probability space $(\Omega, \mathbb{F}, \Pr)$.  Recall that a random variable $F$ is a measurable function $\Omega\mapsto\mathcal{F}$ from the sample space $\Omega$ to an outcome set $\mathcal{F}$.

The probability measure $\Pr:\mathbb{F} \mapsto [0,1]$ assigns likelihoods to the events in the $\sigma-{\rm algebra}$ $\mathbb{F}$.  For any event $E \in \mathbb{F}$, $\Pr(E)$ to denotes the probability of the event.  For events $E,G\in \mathbb{F}$ for which $\Pr(G) > 0$, $\Pr(E|G)$ to denotes the probability of event $E$ conditioned on event $G$.

For realization $z$ of a random variable $Z$, $\Pr(Z=z)$ is a function of $z$.  We denote its value evaluated at $Z$ by $\Pr(Z)$.  Therefore, $\Pr(Z)$ is a random variable (it takes realizations in $[0,1]$ depending on the value of $Z$).  Likewise for realizations $(y,z)$ of random variables $Y,Z$, $\Pr(Z=z|Y=y)$ is a function of $(y,z)$ and $\Pr(Z|Y)$ is a random variable which denotes the value of this function evaluated at $(Y,Z)$.

\section{Problem Formulation}
We begin with the problem formulation for a bandit alignment problem.  We draw inspiration from the hidden-utility bandit \cite{freedman2023active}.
\begin{itemize}
    \item $\Ac_e$ denotes a finite set of environment actions.
    \item $\Ac_h$ denotes a finite set of human query actions.
    \item $\Oc$ denotes a finite set of observations.
    \item $\rho_\phi$ denotes a probability measure $\rho_\phi(o|a)$ (parameterized by $\phi$) for all $o\in\Oc, a \in \Ac_e$.  Hence, $\rho_\phi$ prescribes environment observation probabilities and $\phi:\Omega\mapsto\Phi$ is unknown to the agent.
    \item $\rho_\theta$ denotes a probability measure $\rho_{\theta}(o|a)$ (parameterized by $\theta$) for all $o \in \Oc, a\in \Ac_h$.  Hence, $\rho_\theta$ prescribes human query response probabilities and $\theta:\Omega\mapsto\Theta$ is unknown to the agent.
    \item $\Rc: \Ac\times\Oc\times\Theta\mapsto \R$ denotes a reward function which is a function of $a, o$ and $\theta$.
\end{itemize}
We will use $\Ac$ to denote the union $\Ac_e \cup \Ac_h$
For all $t \in \mathbb{Z}_{+}$, an agent takes an action $A_t \in \Ac$ and receives an observation $O_{t+1}$.  We let $H_t = (A_0, O_1, A_1, O_2, \ldots, O_t)$ denote the agent's \emph{history} of experience through time $t$.  Formally, an agent $\pi$ is a function which maps a history $H_t$ to a probability distribution over $\Ac$.  The agent will also earn reward (or incur cost) $R_{t+1} = \Rc(A_t,O_{t+1}, \theta)$.  However, the agent will never observe $R_{t+1}$.

The agent's goal is to maximize its longrun average reward:
$$\liminf_{T\to\infty}\ \frac{1}{T}\sum_{t=0}^{T-1} \E_{\pi}\left[R_{t+1}\right].$$

While the agent never observes $R_{t+1}$, it can reduce its uncertainty about the reward by querying the human (thereby reducing uncertainty about $\theta$).  Sensible design of bandit alignment problems would attribute \emph{negative} values of $\R_{t+1}$ for actions which query the human, reflecting the associated real world costs.  Meanwhile, actions which interact with the environment should have variable reward which will depend on the observed outcome $O_{t+1}$ and the human's preference $\theta$.  Hence, a bandit alignment problem exhibits the same exploration-exploitation tradeoff as a standard bandit problem, with the additional complexity of also balancing exploration of \emph{human preferences}.  Effective bandit alignment algorithms will \emph{selectively} explore both the environment and the preferences of the human while balancing this with productive output based on the knowledge it has already accrued.

For the purposes of analysis, we consider the following quantity, which we refer to as the \emph{regret} of an agent $\pi$ at horizon $T$:
$$ \Re(\pi,T) = \sum_{t=0}^{T-1} \E_{\pi}\left[R^* - R_{t+1}\right],$$
where $R^* = \argmax_{a\in\Ac} \E[\Rc(A_t,O_{t+1},\theta)|\theta, A_t=a]$ denotes the reward of the \emph{optimal} action.  An agent which minimizes regret as $T\to\infty$ also maximizes the longrun average reward.

The notable difference from a CIRL \cite{hadfield2016cooperative} problem is that $\phi$ is assumed to be \emph{unknown} to the agent.  While in practice an AI agent will never be deployed into the real world with $0$ knowledge of the environment, the same is true for the other extreme: \emph{an AI agent will never be deployed into the real world with $100\%$ knowledge of the environment}.
As we operate in a Bayesian framework, we model this uncertainty at $t=0$ pertaining to the details of the world by letting $\phi:\Omega\mapsto\Phi$ be a \emph{random} variable with prior distribution $\Pr(\theta\in\cdot)$.

Another subtlety is that we do not assume that the human has knowledge of $\phi$.  The human's preferences are dictated by $\theta:\Omega\mapsto\Theta$ which again is a random variable, representing the agent's initial uncertainty about said preferences.  By querying the human, the agent can accumulate observations which reduce its uncertainty about $\theta$ and hence the reward.  Since the agent's goal is to maximize its expected longrun reward, it is in the agent's direct interest to incorporate effective human querying into its overall policy.

\section{Beta-Bernoulli Bandit Alignment Problem}

While we provide a general definition of bandit alignment problems, theoretical and empirical development necessitates thoughtful instantiations.  In this section, we outline a concise yet complete alignment bandit problem modeled after the classic beta-Bernoulli bandit.  By complete, we mean that it exhibits all of the aforementioned tradeoffs.  We believe that this problem is an exciting starting point for theorists and practitioners to ideate on algorithmic solutions to AI alignment.

\subsection{Problem Definition}
The beta-Bernoulli bandit alignment problem is an instance of the bandit alignment problem discussed in the section prior.  Therefore, the problem can be defined by specifying the tuple $(\Ac_e,\Ac_h,\Oc,\rho_{\phi}, \rho_{\theta}, \Rc)$:
\begin{itemize}
    \item $\Ac_e = \{1, 2, \ldots, N\}$
    \item $\Ac_h = \{\bar{1}, \bar{2}, \ldots, \bar{N}\}$
    \item $\Oc = \{0, 1\}$
    \item For all $a, a' \in \Ac_e$, $\phi_a \perp \phi_{a'}$ if $a \neq a'$,\\
    $\Pr(\phi_a\in\cdot) = {\rm beta}(1,1)$,\\
    and $\rho_\phi(\cdot|a) = {\rm bernoulli}(\phi_a)$.
    \item For all $a, a' \in \Ac_h$, $\theta_a, \perp \theta_{a'}$ if $a \neq a'$,\\
    $\Pr(\theta_a\in\cdot) = {\rm beta}(1, 1)$,\\
    and $\rho_{\theta}(\cdot|a) = {\rm bernoulli}(\theta_a)$.
    \item $\Rc(a,o,\theta) = $
    \begin{align*}
        \begin{cases}
            -1 & \text{ if } a \in \Ac_{h}\\
            o\cdot\theta_{\bar{a}} + (1-o)\cdot(1-\theta_{\bar{a}}) & \text{ if } a \in \Ac_{e}\\
        \end{cases}.
    \end{align*}
\end{itemize}
\begin{figure}[H]
    \centering
    \includegraphics[width=0.9\linewidth]{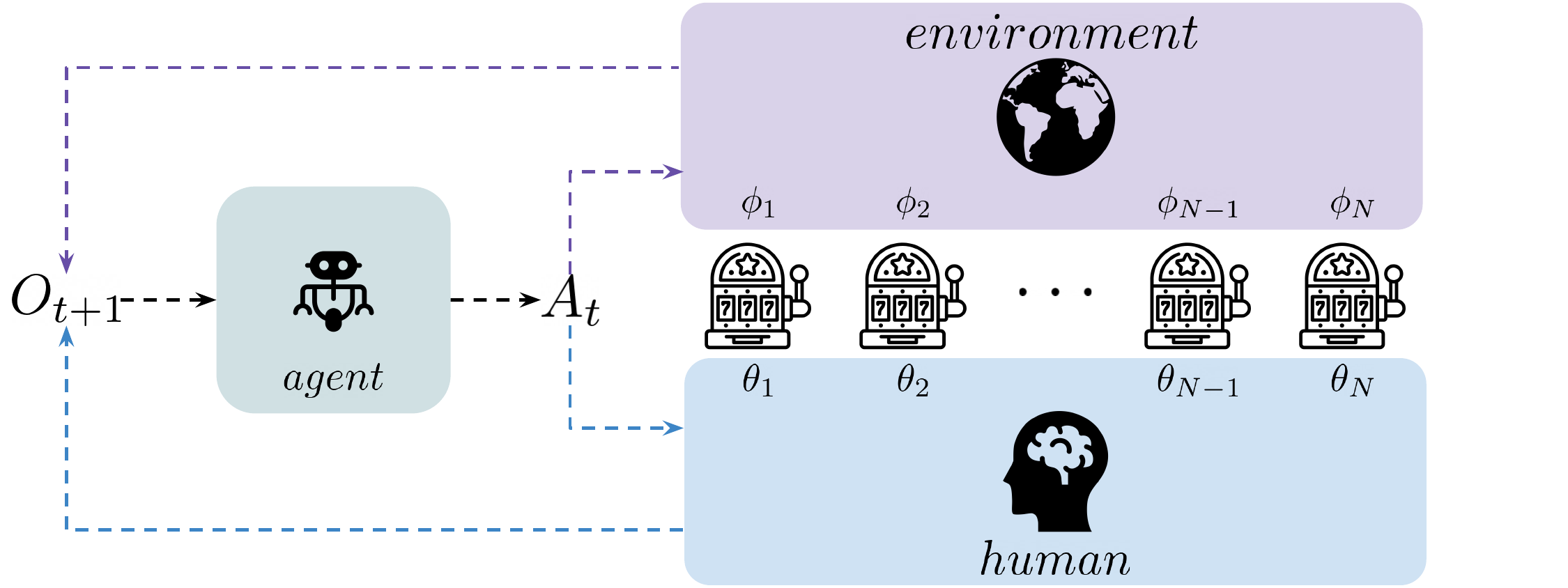}
    \caption{The above diagram depicts the beta-Bernoulli bandit alignment problem.  Each bandit arm has a probability $\phi_i \in [0,1]$ such that $\Pr(O_{t+1}=1|\phi_i, A_t\in\Ac_e) = \phi_i$.  Meanwhile, the human exhibits a preference $\theta_i \in [0,1]$ for each of the arms which impacts the mean (unobserved) reward of said arm.  For all $t$, the agent selects an action $A_t$ which either interfaces with the environment or the human and receives a corresponding observation $O_{t+1}$.  The reward is $-1$ if the agent queries the human and unobserved (but dependent on $O_{t+1}, \theta_1$) if the agent interacts with the environment.}
    \label{fig:betaBernoulliBanditAlign}
\end{figure}
\subsection{Remarks}

We begin with a high-level description of the problem.  The problem exhibits $N$ bandit arms represented by $\Ac_e$, each with a beta-Bernoulli outcome.  Associated with each bandit arm $a$ is a query $\bar{a}$ which provides information pertaining to the human's preference $\theta_{\bar{a}}$ of said arm.  We now provide some intuition for the reward function.  Note that 
\begin{align*}
    &\ \E[\Rc(A_t,O_{t+1},\theta)|\phi, \theta, A_t=a]\\
    & = [\phi_{a}, 1-\phi_a]^\top [\theta_{\bar{a}}, 1-\theta_{\bar{a}}].
\end{align*}
Therefore, the action which maximizes reward is the one for which the environment $[\phi_a, 1-\phi_a]$ outcomes and the human preferences $[\theta_{\bar{a}}, 1-\theta_{\bar{a}}]$ are most \emph{aligned}.  We do not make any claims as to the reasonability of this reward function beyond the scope of this problem.  We simply use it as sufficient example to elucidate the inherent challenges present in aligning AI agents.

For the standard beta-Bernoulli bandit problem, an optimal (and tractable) algorithm exists via Gittins indices \citep{gittins1979bandit}.  However, the beta-Bernoulli bandit alignment problem does not satisfy the conditions necessary for tractable computation of Gittins indices.  Furthermore, since we hope for this theory to extend to complex problem instances involving neural networks, we restrict our attention to algorithms for which there exist scaleable extensions for neural networks.  However, a thorough analysis of optimal performance in the beta-Bernoulli bandit problem to establish lower bounds is an interesting direction for future theoretical work.

A subtlety which becomes apparent after some thought is that an agent must not \emph{over-explore}.  Over-exploration can occur across two axes in this problem: $1)$ over-querying the human, $2)$ over-exploring the environment.  $1)$ manifests as a result of the \emph{negative} reward of $-1$ associated with any action which queries a human.  Any algorithm which queries the human \emph{periodically} with fixed period $\tau$ will incur \emph{linear regret}.

\begin{lemma}
    For all $T \in\mathbb{Z}_{+}$, $\tau < T$ and $\delta > 0$, if $\pi$ is such that for all $i \leq \lfloor T/\tau\rfloor$,
    $$\Pr( \cup_{t \in [i\tau, (i+1) \tau]}\  (A_t \in \Ac_h)) \geq \delta,$$
    then
    $$\Re(\pi, T) \geq \frac{\delta}{\tau} T.$$
\end{lemma}

In the standard beta-Bernoulli bandit problem, algorithms which suffer $\Omega(T)$ regret are consigned as ineffective as their per-timestep regret averaged across time does \emph{not} diminish to $0$ even as $T \to \infty$.  Effective algorithms for the beta-Bernoulli bandit problem typically exhibit $O(\sqrt{|\Ac|T})$ regret \citep{russo2016information}.  

$2)$ manifests as a natural consequence of being a bandit problem.  Over-exploration of unpromising actions will also lead to unfavorable regret:

\begin{lemma}
    For all $T \in \mathbb{Z}_{+}$, if $\pi$ is such that there exists $\epsilon > 0$ s.t. for all $t < T$ and $a \in \Ac$, $\Pr(A_t=a) \geq \epsilon$, then
    $$\Re(\pi,T) = \Omega(T).$$
\end{lemma}

This result consigns any algorithm which performs uniform exploration such as epsilon-greedy as ineffective.  This is intuitive as it is irrational to continuing to explore actions at a fixed rate even after they have demonstrated to be unfruitful.  In the following section, we will demonstrate how naive exploration algorithms and even touted algorithms from the bandit literature (Thompson sampling) fail to deliver effective solutions to the bandit alignment problem.

The final aspect of this environment which we outline is that fully identifying the environment requires infinite information: $\H(\theta,\phi) = \infty$.  This is because $\theta, \phi$ are continuous random variables and each observation only conveys a finite number of bits about $\theta, \phi$.  This detail is crucial in that it captures the \emph{limitless} complexity of the world and the \emph{limited} granularity/veracity in which humans and machines can interpret it.  In environments which lack this property, an agent may ``finish exploring'' i.e. fully identify the environment after a finite time horizon, a preposterous assumption when attempting to provide insight for aligning superintelligent systems.  As we will see in the following section, this detail produces significant ramifications for naive solutions to our bandit alignment problem.

\section{Insufficiency of Standard Methods}
In this section, we will demonstrate theoretically that existing algorithms prevalent in the literature fail to provide sufficient solutions to the beta-Bernoulli bandit alignment problem.  A solution is insufficient if its regret is $\Omega(T)$ or $\Omega(|\Ac|)$.  We begin with a class of naive exploration algorithms referred to as ``explore then exploit'' and then move onto Thompson sampling, an effective algorithm in the standard beta-Bernoulli bandit environment.

\subsection{Explore then Exploit}
For all $t$, an agent $\pi$ must decide which action $a$ to take based off the information it has already accrued $H_t$.  The \emph{reward-greedy} agent is one which \emph{greedily} optimizes reward based on $H_t$:
$$A_t = \argmax_{a\in\Ac}\ \E[R_{t+1}|H_t,A_t=a].$$
However, it is easy to see that such algorithms are susceptible to self-perpetuating feedback loops of only repeatedly selecting the actions for which it already has accrued information about.  Meanwhile, the \emph{information-greedy} agent is one which acts \emph{greedily} optimizes information acquisition base on $H_t$:
$$A_t = \argmax_{a\in\Ac}\ \I(O_{t+1};\theta,\phi|H_t=H_t, A_t=a).$$
Paradigms such as A/B testing, or apprenticeship learning design explicit information-greedy phase followed by a reward-greedy phase.  We group these methods under the umbrella of ``explore then exploit'' agents.  

An ``explore then exploit'' agent is identified by $\tau$ such that for 
$t \leq \tau$, $\pi$ is the information-greedy agent and for $t > \tau$, $\pi$ is the reward-greedy agent.  However, we have the following result which demonstrates

\begin{theorem}{\bf (regret bound)}\label{th:explore_exploit}
    If $\pi$ is an ``explore then exploit'' agent, then
    $$\Re(\pi, T) = \Omega(|\Ac| + T).$$    
\end{theorem}

We now provide some intuition regarding Theorem \ref{th:explore_exploit}.  In the exploration phase, the information-greedy agent will select the action $a\in \Ac$ which has been selected the \emph{fewest} number of times so far in $H_t$.  This explicit uniform exploration scheme will result in the $\Omega(|\Ac|)$ term since in order to accrue sufficient information about the reward of the optimal action, the agent must explore all other actions equally as well.

Meanwhile in the exploitation phase, no matter how large $\tau$ is, there exists a nonzero probability (dependent on $\tau$) that the optimal action was not identified during exploration.  In this scenario, the agent which exploits its information will incur \emph{linear} regret in $T$, resulting in the $\Omega(T)$ term in the bound.  Recall that linear regret in $T$ implies that the agent does not reliably identify the best action even as the number of interactions goes to $\infty$.  While we establish this result in a stylized problem, it brings to light some serious concerns of applying these ``explore then exploit'' methods to aligning superintelligence.

``Explore then exploit'' agents exhibit this pitfall due to the fact that $1)$ they explore in a reward-insensitive manner and $2)$ they only explore finitely often.  Eventually, we will ameliorate these issues with an algorithm which explores infinitely often in a reward-\emph{sensitive} manner.  However, we will first analyze the performance of a popular algorithm from the bandit literature which is touted for its exploration capabilities: Thompson sampling.

\subsection{Thompson Sampling}

In the standard beta-Bernoulli bandit, Thompson sampling (TS) is an efficient algorithm has offers strong theoretical guarantees.  In that problem setting, TS achieves regret which is $O(\sqrt{T|\Ac|}$, i.e. sublinear in both the horizon $T$ and the size of the action set $|\Ac|$.  Therefore, it is a natural algorithmic choice for our beta-Bernoulli bandit \emph{alignment} problem.
We demonstrate that despite its merits in the standard bandit environments, TS produces incoherent behavior in bandit alignment problems.

We first outline the Thompson sampling agent, which we denote by $\pi_{\rm ts}$.  For all $t \in \mathbb{Z}_{++}$, 
$$\pi_{ts}(\cdot|H_t)\ =\ \Pr(A^*=a|H_t),$$
where $A^* = \argmax_{a\in\Ac} \E[R_{t+1}|\theta,\phi,A_t=a]$.  However, we notice that for any action $\bar{a}\in \Ac_h$,
$$\Pr(A^*=\bar{a}|H_t) \overset{a.s.}{=} 0.$$
This is because actions which query the human incur a reward of $-1$ whereas all environment actions have non-negative expected reward.  As a result, TS will \emph{never} query the human.  Some simple calculations provide the following result.

\begin{lemma}\label{le:ts_actions}
    For all $t \in \mathbb{Z}_{+}$, let $\pi_{\rm ts}^{(t)}(\cdot)$ denote the Thompson sampling agent at time $t$, then
    $$\pi_{\rm ts}^{(t)}(\cdot)\ \overset{a.s.}{=}\ \begin{cases}
        \frac{1}{|\Ac_e|}\ & \text{ if } \cdot \in \Ac_e\\
        0 & \text{ if } \cdot \in \Ac_h\\
    \end{cases}.$$
\end{lemma}

Lemma \ref{le:ts_actions} establishes that for all $t \in \mathbb{Z}_{+}$, the TS agent selects actions uniformly at random from $\Ac_e$.  As a result, this algorithm incurs linear regret even as $T \to \infty$.

\begin{theorem}{\bf (ts regret bound)}
    $$\Re(\pi_{\rm ts}, T) = \Omega(T).$$
\end{theorem}

The issue with Thompson Sampling (and other common bandit algorithms such as UCB) is that they only sample an action if it is statistically plausible that it is \emph{optimal}.  So while these algorithms are reward-sensitive, they are \emph{information-blind}.  In the following section, we will analyze information directed sampling, an algorithm which overcomes these hurdles by being simultaneously reward and information sensitive.  

\section{Information Directed Sampling}
We have demonstrated that Thompson sampling fails to demonstrate any useful behavior in the beta-Bernoulli bandit alignment problem.  However, in this section, we demonstrate that Information-Directed Sampling (IDS) \citep{russo2014learning} \emph{does} produce favorable theoretical regret guarantees.  We begin by outlining the IDS algorithm.

\subsection{Algorithm}
We begin by defining the conditional information ratio.  For all $t \in \mathbb{Z}_{+}$, let the conditional information ratio of an agent $\pi$ at time $t$ be:
$$\Gamma_t(\pi)\ =\ \frac{\E_\pi\left[R^*-R_{t+1}|H_t\right]^2}{\I(\theta,\phi;A_t, O_{t+1}|H_t=H_t)},$$
where $A_t \sim \pi(\cdot|H_t)$ (hence the $\pi$ in the subscript of the expectation).  Note that the $\Gamma_t(\pi)$ is a random variable (since $H_t$ is a random variable).  The choice of agent $\pi$ will dramatically impact the information ratio.  In general, we can upper bound the regret of any agent $\pi$ via its conditional information ratio:

\begin{restatable}{theorem}{regretir}{\bf (information ratio regret bound)}\label{th:regret_ir}
    For all $T \in \mathbb{Z}_{++}$ and algorithms $\pi$,
    $$\Re(\pi, T) \ \leq\ \sqrt{\sum_{t=0}^{T-1}\E_{\pi}\left[\Gamma_t(\pi)\right]} \cdot \sqrt{\I(\theta,\phi;H_T)}.$$
\end{restatable}

Theorem \ref{th:regret_ir} establishes that an algorithm with a controlled information ratio will also exhibit a controlled regret.  Therefore a sensible algorithmic design would be to \emph{minimize} the information ratio.  This results in the IDS algorithm: for all $t \in \mathbb{Z}_{+}$,
$$\pi_{\rm ids}(\cdot|H_t) = \argmin_{\pi\in \Dc(\Ac)}\ \Gamma_t(\pi),$$
where $\Dc(\Ac)$ denotes the set of probability mass functions on $\Ac$.  \cite{russo2014learning} demonstrate that the above optimization is \emph{convex}.  Moreover, there exists an optimal solution which randomizes over at most $2$ actions.  In the following section we provide a concrete regret bound for $\pi_{\rm ids}$ for the beta-bernoulli bandit alignment problem.
 
\subsection{Regret Bound}

By Theorem \ref{th:regret_ir}, upper bounds on the sum of expected information ratios and the mutual information $\I(\theta,\phi;H_T)$ provide an upper bound on regret.  In this section, we derive a regret upper bound for IDS via upper bounding the two aforementioned quantities.  Proofs can be found in the appendix.  We begin with the information ratio:

\begin{lemma}{\bf (ids information ratio bound)}\label{le:info_ratio}
    For all $T \in \mathbb{Z}_{+}$,
    $$\sum_{t=0}^{T-1}\ \E_{\pi_{\rm ids}}\left[\Gamma_t(\pi_{\rm ids})\right]\ \leq\ 33\sqrt{|\Ac|T^3}.$$
\end{lemma}

We now provide an upper bound on the mutual information.

\begin{lemma}{\bf (mutual information bound)}\label{le:mi}
    For all $T\in \mathbb{Z}_{++},$
    $$\I(H_T;\theta,\phi) \ \leq\ |A|\ln(4T).$$
\end{lemma}

With these in place, we present the main result.

\begin{restatable}{theorem}{idsRegret}{\bf (ids regret bound)}\label{th:ids}
    For all $T \in \mathbb{Z}_{+}$,  
    $$\Re(\pi_{\rm ids},T)\ \leq\ \sqrt{33\ln (4T)}|\Ac|^{\frac{3}{4}}T^{\frac{3}{4}}.$$
\end{restatable}

Theorem \ref{th:ids} follows directly from Theorem \ref{th:regret_ir} and Lemmas \ref{le:info_ratio} and \ref{le:mi}.  Notably, the regret upper bound is $O(|\Ac|^{3/4}T^{3/4})$ ignoring log factors, hence \emph{sublinear} in $|\Ac|$ \emph{and} $T$.  By seeking information in a reward-sensitive manner, IDS is able to escape the pitfalls of ``explore then exploit'' agents and TS which only prioritize at most one of the two at a time.

In the standard beta-Bernoulli bandit environment, both TS and IDS achieve regret bounds which are $O(\sqrt{|\Ac|T})$.  Therefore, there is a considerable gap between the upper bounds for the standard bandit and bandit alignment problems.  We conjecture that IDS can actually match the $O(\sqrt{|\Ac|T})$ performance and therefore the gap can be reduced via improved analytic techniques.  The empirical results of the following section appear to corroborate this conjecture.

\section{Empirical Results}

In this section, we provide empirical results which compare IDS and ``explore then exploit'' agents in the beta-Bernoulli bandit alignment problem.  The results corroborate our theoretical analysis and as mentioned, IDS appears to perform significantly \emph{better} than what is suggested by Theorem \ref{th:ids}.

\subsection{Experimental Details}
\begin{figure}[H]
    \centering
    \includegraphics[width=0.5\linewidth]{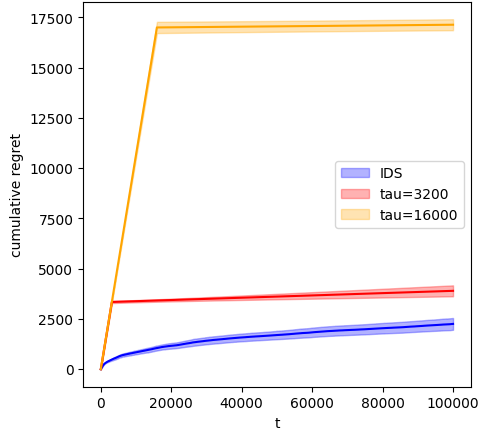}
    \caption{The above plot depicts the performance of IDS and the ``explore then exploit'' agent for $2$ different values of $\tau$.  Evidently, the regret accumulated in the reward-blind exploration phase is very costly.  Furthermore, figure \ref{fig:exploreExploit} demonstrates that even with this exploration, the ``explore then exploit'' agents were unable to identify the best action. The shaded area denotes $\pm$ standard error.}
    \label{fig:regretComparison}
\end{figure}

We evaluated IDS and two ``explore then exploit'' algorithms $(\tau = 3200, 16000)$.  We selected an environment with $|\Ac_e| = 16$ arms and averaged the cumulative regret of each method across $10$ random problem instances (with each $\theta_{\bar{a}}, \phi_a$ drawn randomly from the beta prior).  The results if this experiment are presented in figure \ref{fig:regretComparison}.

In the following section, we provide a detailed analysis of our empirical findings.

\subsection{Discussion}

\begin{figure}[H]
    \centering
    \includegraphics[width=0.4\linewidth]{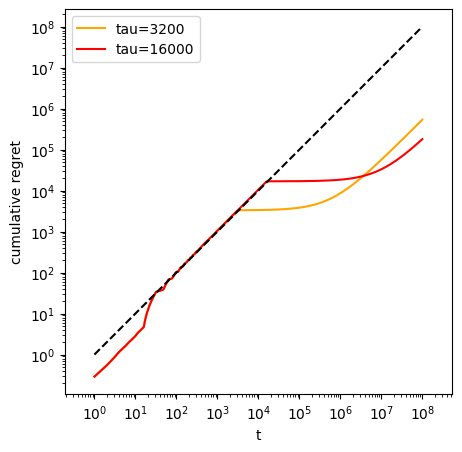}
    \caption{The above plot depicts the cumulative regret of ``explore then exploit'' agents for $\tau = 3200, 16000$.  Note that both axes are in logarithmic scale and hence the dotted line depicts $2t$, exemplified by the slope of $1$.  We notice that in the exploration phase, the regret is linear (since it is also slope $1$) and in the same insues in the exploitation phase (slope increase up to $1$ again as $t$ increases).}
    \label{fig:exploreExploit}
\end{figure}

Figure \ref{fig:regretComparison} demonstrates that as suggested by Theorem \ref{th:explore_exploit}, the regret of ``explore then exploit'' agents grows \emph{linearly} in the horizon.  This indicates that even when each action is explored $100$ or $500$ times (for $\tau = 3200, 16000$ respectively), there is still a significant number of instances in which the optimal action is not identified.  Furthermore, significant cost is incurred in the process of exploring the environment in a reward-blind fashion.  As apparent in figure \ref{fig:exploreExploit},  even after the exploration phase, the agent is unable to identify the best action.  The figure plots the regret of each agent vs $t$ with both axes in logarithmic scale.  The reference dotted line depicts $t$ regret as the slope is $1$.  Evidently, the performance of the ``explore then exploit'' agents is $\Omega(t)$ which indicates that the the performance gap between these agents and the IDS agent will continue to widen as $t$ grows.

\begin{figure}[H]
    \centering
    \includegraphics[width=0.4\linewidth]{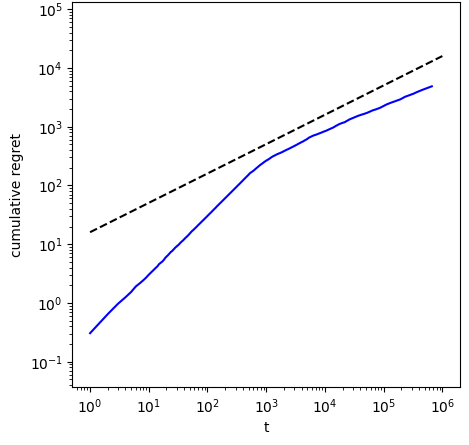}
    \caption{The above plot depicts the cumulative regret of IDS.  Note that both axes are in logarithmic scale and hence the dotted line depicts $16\sqrt{t}$, exemplified by the slope of $1/2$.  The plot suggests that the cumulative regret of IDS is $O(\sqrt{t})$.}
    \label{fig:idsLogRegret}
\end{figure}

Meanwhile the regret of IDS is evidently \emph{sublinear} in the horizon and hence the performance gap will continue to widen as $t \to \infty$.  This point is made clear by figure \ref{fig:idsLogRegret} which plots the cumulative regret of ids with both axes in logarithmic scale.  The reference dotted line depicts $\sqrt{t}$ regret as the slope is $0.5$.  Evidently, for large enough $t$, the slope of the average cumulative regret appears to be close to $0.5$, indicating that the cumulative regret of IDS is $O(\sqrt{t})$.  This is tighter than Theorem \ref{th:ids} which characterizes the regret of IDS as $O(t^{3/4})$.  This performance matches the theoretical and empirical results in the standard beta-Bernoulli bandit problem \citep{russo2014learning}.  However, in the beta-Bernoulli bandit problem, algorithms such as TS were sufficient to obtaining $O(\sqrt{t})$ regret.  In the beta-Bernoulli bandit \emph{alignment} problem, this performance is only afforded to IDS as it effectively seeks information in a reward-sensitive manner.  It remains an open theoretical question as to how to tighten the results of Theorem \ref{th:ids} to match those of empirical evaluation. 

\section{Conclusion}
In this paper we elucidate some of the challenges we will encounter in our efforts to align AI systems.  Some of these challenges including simultaneous exploration of the environment and human preferences and accounting of cumulative reward/regret had not been addressed prior to this work.  We provide a simple yet complete problem instance which we believe can serve as both a testbed for theoretical and empirical research in AI alignment.  Theoretical questions pertaining to tightening the results of Theorem \ref{th:ids} and empirical questions about discovering other effective algorithms serve as fascinating directions for future research.


\bibliography{references}  

\newpage
\appendix
\section{Proof of General Regret Bound}
We now establish a regret bound in terms of the information ratio.  Recall the conditional information ratio:
$$\Gamma_t\ =\ \frac{\E_\pi\left[R_t^*-R_t|H_t\right]^2}{\I(\theta,\phi;(A_t, O_{t,A_t})|H_t=H_t)}.$$
Then, let the unconditional information ratio be:
$$\bar{\Gamma}_t\ = \ \frac{\E_\pi\left[R_t^*-R_t\right]^2}{\I(\theta,\phi;(A_t, O_{t,A_t})|H_t)}.$$
The following result bounds the regret of an agent $\pi$ via its conditional information ratio:
\regretir*
\begin{proof}
    Let $\bar{\Gamma}_t = \E_{\pi}[\Gamma_t(\pi)]$.  Then
    \begin{align*}
        \sum_{t=0}^{T-1} \E_\pi\left[R^*_t - R_t\right]
        & = \sum_{t=0}^{T-1} \sqrt{\left(\E_\pi\left[R^*_t - R_t\right]\right)^2}\\
        & = \sum_{t=0}^{T-1} \sqrt{\bar{\Gamma}_t \cdot \I(\theta,\phi;A_t, O_{t,A_t}|H_t)}\\
        & \overset{(a)}{\leq} \sqrt{\sum_{t=0}^{T-1}\ \bar{\Gamma}_t} \cdot \sqrt{\sum_{t=0}^{T-1} \I(\theta,\phi;A_t,O_{t,A_t}|H_t)}\\
        & \overset{(b)}{=} \sqrt{\sum_{t=0}^{T-1}\ \bar{\Gamma}_t} \cdot \sqrt{\I(\theta,\phi;H_T)}\\
        & \overset{(c)}{\leq} \sqrt{\sum_{t=0}^{T-1}\ \frac{\E_\pi\left[\E_{\pi}\left[R^*_t-R_t|H_t\right]^2\right]}{\E_\pi\left[ \I(\theta,\phi;A_t, O_{t,A_t}|H_t=H_t) \right]}} \cdot \sqrt{\I(\theta,\phi;H_T)}\\
        & \overset{(d)}{\leq} \sqrt{\sum_{t=0}^{T-1}\ \E_\pi\left[\frac{\E_{\pi}\left[R^*_t-R_t|H_t\right]^2}{ \I(\theta,\phi;A_t, Y_{t,A_t}|H_t=H_t)}\right]} \cdot \sqrt{\I(\theta,\phi;H_T)}\\
        & = \sqrt{\sum_{t=0}^{T-1}\ \E_\pi\left[\Gamma_t\right]} \cdot \sqrt{\I(\theta,\phi;H_T)}\\
    \end{align*}
    where $(a)$ follows from Cauchy Bunyakovsky Schwarz, $(b)$ follows from the chain rule of mutual information, $(c)$ follows from the fact that for any real-valued random variable $X$, $\E[X]^2 \leq \E[X^2]$, and $(d)$ follows from Jensen's inequality and the fact that quadratic over linear is convex.
\end{proof}

\section{Proof of Information-Theoretic Quantities}

We now provide results which upper and lower bound the mutual information of observations and $\theta,\phi$ for our beta-Bernoulli bandit alignment problem.  We start with this general result for beta and bernoulli distributed random variables:

\subsection{Mutual Information Bounds for Beta-Bernoulli Random Variables}

\begin{lemma}\label{le:mutual_info_bound}
    For all $\alpha, \beta, t \in \mathbb{Z}_{++}$, if $\Pr(\theta\in\cdot) = {\rm Beta}(\alpha,\beta)$ and $X \sim {\rm Bernoulli}(\theta)$, then
    $$\frac{1}{4(\alpha+\beta)}\ \leq\ \I(X;\theta)\ \leq\ \frac{1}{2(\alpha+\beta)}.$$
\end{lemma}
\begin{proof}
    \begin{align*}
        \I(X; \theta)
        & = \diffentropy(\theta) - \diffentropy(\theta|X)\\
        & = \ln \frac{\Gamma(\alpha)\Gamma(\beta)}{\Gamma(\alpha+\beta)} - \left(\alpha-1\right)\psi(\alpha) - (\beta-1)\psi(\beta) + (\alpha + \beta -2)\psi(\alpha + \beta)\\
        &\ - \frac{\alpha}{\alpha+\beta} \left[\ln \frac{\Gamma(\alpha+1)\Gamma(\beta)}{\Gamma(\alpha+\beta+1)} - \left(\alpha\right)\psi(\alpha+1) - (\beta-1)\psi(\beta) + (\alpha + \beta -1)\psi(\alpha + \beta+1)\right]\\
        &\ - \frac{\beta}{\alpha+\beta} \left[\ln \frac{\Gamma(\alpha)\Gamma(\beta+1)}{\Gamma(\alpha+\beta+1)} - \left(\alpha-1\right)\psi(\alpha) - (\beta)\psi(\beta+1) + (\alpha + \beta -1)\psi(\alpha + \beta+1)\right]\\
        & = \ln (\alpha + \beta) + \ln \Gamma(a) \Gamma(\beta) - \frac{\alpha}{\alpha+\beta}\ln \Gamma(\alpha+1)\Gamma(\beta) - \frac{\beta}{\alpha + \beta}\ln \Gamma(\alpha)\Gamma(\beta+1)\\
        &\ + (\alpha+\beta-2)\psi(\alpha+\beta) - (\alpha+\beta-1)\psi(\alpha+\beta+1)\\
        &\ -(\alpha-1)\psi(\alpha) + \frac{\alpha}{\alpha+\beta}\alpha\left(\psi(\alpha)+\frac{1}{\alpha}\right) + \frac{\beta}{\alpha+\beta}(\alpha-1)\psi(\alpha)\\
        &\ -(\beta-1)\phi(\beta) + \frac{\beta}{\alpha+\beta}\beta\left(\psi(\beta)+\frac{1}{\beta}\right) + \frac{\alpha}{\alpha+\beta}(\beta-1)\psi(\beta)\\
        & = \ln(\alpha+\beta) - \frac{\alpha}{\alpha+\beta}\ln\alpha - \frac{\beta}{\alpha+\beta}\ln\beta\\
        &\ + (\alpha+\beta-2)\psi(\alpha+\beta) - (\alpha+\beta-1)\left(\psi(\alpha+\beta)+\frac{1}{\alpha+\beta}\right)\\
        &\ + \frac{\alpha}{\alpha+\beta}\psi(\alpha) + \frac{\beta}{\alpha+\beta}\psi(\beta) + 1\\
        & = \ln(\alpha+\beta) - \frac{\alpha}{\alpha+\beta}\ln\alpha - \frac{\beta}{\alpha+\beta}\ln\beta\\
        &\ - \psi(\alpha+\beta) - \frac{\alpha+\beta-1}{\alpha+\beta} + \frac{\alpha}{\alpha+\beta}\psi(\alpha) + \frac{\beta}{\alpha+\beta}\psi(\beta) + 1\\
        & = \frac{1}{\alpha+\beta} + \ln(\alpha+\beta) - \psi(\alpha+\beta) + \frac{\alpha}{\alpha+\beta}\left(\psi(\alpha) - \ln\alpha\right) + \frac{\beta}{\alpha+\beta}(\psi(\beta)-\ln\beta)
    \end{align*}
    We begin with the lower bound:
    \begin{align*}
        &\ \frac{1}{\alpha+\beta} + \ln(\alpha+\beta) - \psi(\alpha+\beta) + \frac{\alpha}{\alpha+\beta}\left(\psi(\alpha) - \ln\alpha\right) + \frac{\beta}{\alpha+\beta}(\psi(\beta)-\ln\beta)\\
        &\overset{(a)}{\leq} \frac{1}{\alpha+\beta} + \ln(\alpha+\beta) - \psi(\alpha+\beta) + \frac{1}{2}\left(\psi\left(\frac{\alpha+\beta}{2}\right) - \ln\frac{\alpha+\beta}{2}\right) + \frac{1}{2}\left(\psi\left(\frac{\alpha+\beta}{2}\right)-\ln\frac{\alpha+\beta}{2}\right)\\
        & = \frac{1}{\alpha+\beta} + \ln(\alpha+\beta) - \psi(\alpha+\beta) + \phi\left(\frac{\alpha+\beta}{2}\right) - \ln\frac{\alpha+\beta}{2}\\
        & \overset{(b)}{\leq} \frac{1}{2(\alpha+\beta)},
    \end{align*}
    where $(a)$ follows from the fact that the expression is maximized for $\alpha=\beta=(\alpha+\beta)/2$ and $(b)$ can be shown in the limit as $\alpha+\beta \to\infty$ and numerically for small $\alpha+\beta$. 
 The lower bound can be demonstrated numerically.  The result follows.
    \begin{align*}
        \I(O_{a}; \theta, \phi|H_t=H_t, A_t=a)
        & = \diffentropy(\theta, \phi|H_t=H_t, A_t=a) - \diffentropy(\theta, \phi|H_t=H_t, A_{t}=a, O_{a})\\
    \end{align*}
\end{proof}

By applying Lemma \ref{le:mutual_info_bound} to our problem setting, we have the following:

\begin{corollary}\label{cor:mutualinfo}
    For all $t\in \mathbb{Z}_{+}$ and $a \in \Ac$, the following holds almost surely:
    $$\frac{1}{4n_{a,t}}\ \leq\ \I(O_{t+1};\theta,\phi|H_t=H_t,A_t=a)\ \leq\ \frac{1}{2n_{a,t}}$$
\end{corollary}
\begin{proof}
    The result follows directly from Lemma \ref{le:mutual_info_bound}.
\end{proof}

\subsection{Bounding Environment and Human Information}

In Theorem \ref{th:regret_ir}, the second term on the RHS cannot simply be upper bounded by $\H(\theta,\phi)$ since $\theta$ and $\phi$ are continuous random variables.  As a result, we have the following series of results which upper bound $\I(H_T;\theta,\phi)$ via rate-distortion theory as in \cite{jeon2024informationtheoreticfoundationsmachinelearning}.  The following result defines a particular lossy compression of $\theta,\phi$ and bounds its distortion.

\begin{lemma}\label{le:info_env}
    For all $t \in \mathbb{Z}_{++}$ and $ \epsilon \in [0, 0.25]$, if $S_\epsilon$ is an $\epsilon$-cover of the interval $[0,1]$ w.r.t the L1 norm, if for all $a\in \Ac_e, \bar{a}\in \Ac_h$,
    $$\tilde{\theta}_{\bar{a}} = \begin{cases}
        \min_{\nu \in \{s \in S_\epsilon: \theta_{\bar{a}} \leq s\leq 0.5\}} \nu & \text{ if } \theta_{\bar{a}} \leq 0.5\\
        \max_{\nu \in \{s \in S_\epsilon: 0.5 \leq s\leq \theta_{\bar{a}}\}} \nu & \text{ if } \theta_{\bar{a}} \geq 0.5\\
    \end{cases},\quad
    \tilde{\phi}_a = \begin{cases}
        \min_{\nu \in \{s \in S_\epsilon: \phi_{a} \leq s\leq 0.5\}} \nu & \text{ if } \phi_a \leq 0.5\\
        \max_{\nu \in \{s \in S_\epsilon: 0.5 \leq s\leq \phi_a\}} \nu & \text{ if } \phi_a \geq 0.5\\
    \end{cases},$$
    then
    $$\I(O_{t+1},A_t;\theta,\phi|\tilde{\theta},\tilde{\phi}, H_t) \ \leq\ 2 \epsilon^2 \ln \frac{1}{\epsilon^2}.$$
\end{lemma}
\begin{proof}
    \begin{align*}
        \I(O_{t+1},A_t;\theta,\phi|\tilde{\theta},\tilde{\phi}, H_t)
        & \leq \I(O_{t+1},A_t;\theta,\phi|\tilde{\theta},\tilde{\phi})\\
        & = \I(O_{t+1};\theta,\phi|\tilde{\theta},\tilde{\phi}, A_t)\\
        & \leq \I(O_{t+1};\theta,\phi|\theta\leftarrow\tilde{\theta},\tilde{\theta}\leftarrow\tilde{\phi}, A_t)\\
        & = \I(O_{t+1};\phi_a|\phi_a\leftarrow\tilde{\phi}_a, A_t=a)\\
        & = \E\left[-\phi_a \ln \frac{\tilde{\phi}_a}{\phi_a} - (1-\phi_a) \ln \frac{1-\tilde{\phi}_a}{1-\phi_a}\right]\\
        & \overset{(a)}{=} \E\left[-\phi_a \ln \frac{\tilde{\phi}_a}{\phi_a} - (1-\phi_a) \ln \frac{1-\tilde{\phi}_a}{1-\phi_a}\bigg|\phi_a \geq 0.5\right]\\
        & = \E\left[-\phi_a \ln \frac{\phi_a + (\tilde{\phi}_a-\phi_a)}{\phi_a} - (1-\phi_a) \ln \frac{1-\phi_a + (\phi_a - \tilde{\phi}_a)}{1-\phi_a}\bigg|\phi_a \geq 0.5\right]\\
        & \leq \E\left[-\phi_a \ln \left(1-\frac{\epsilon}{\phi_a}\right) - (1-\phi_a) \ln \left(1 +\frac{\epsilon}{1-\phi_a}\right)\bigg|\phi_a \geq 0.5\right]\\
        & \overset{(b)}{\leq} \E\left[\frac{\epsilon}{1-\frac{\epsilon}{\phi_a}} - \frac{\epsilon}{1 + \frac{\epsilon}{1-\phi_a}}\bigg|\phi_a \geq 0.5\right]\\
        & = \int_{0.5}^{1.0} 2.0 \cdot \frac{\epsilon^2}{(1-\phi_a+\epsilon)(\phi_a-\epsilon)}\ d\phi_a\\
        & = 2 \epsilon^2 \left(\ln \frac{1-\epsilon}{\epsilon} - \ln \frac{0.5-\epsilon}{0.5+\epsilon}\right)\\
        & \overset{(c)}{\leq} 2 \epsilon^2 \ln \frac{1}{\epsilon^2},
    \end{align*}
    where $(a)$ follows from symmetry of a ${\beta}(1,1)$ distributed random variable, $(b)$ follows from the fact that $\ln(x) \geq 1 - 1/x$, and $(c)$ follows from the fact that $\epsilon \leq 0.25$.
\end{proof}

We can now upper bound the mutual information $\I(\theta,\phi;H_T)$ via standard techniques involving rate-distortion theory.

\begin{lemma}
    For all $T \in \mathbb{Z}_{+}$, 
    $$\I(\theta,\phi;H_T)\ \leq\ |\Ac|\ln (4T)$$
\end{lemma}
\begin{proof}
    For any $ \epsilon \in [0, 0.25]$, if $S_\epsilon$ is an $\epsilon$-cover of the interval $[0,1]$ w.r.t the L1 norm, let for all $a\in \Ac_e, \bar{a}\in \Ac_h$,
    $$\tilde{\theta}_{\bar{a}} = \begin{cases}
        \min_{\nu \in \{s \in S_\epsilon: \theta_{\bar{a}} \leq s\leq 0.5\}} \nu & \text{ if } \theta_{\bar{a}} \leq 0.5\\
        \max_{\nu \in \{s \in S_\epsilon: 0.5 \leq s\leq \theta_{\bar{a}}\}} \nu & \text{ if } \theta_{\bar{a}} \geq 0.5\\
    \end{cases},\quad
    \tilde{\phi}_a = \begin{cases}
        \min_{\nu \in \{s \in S_\epsilon: \phi_{a} \leq s\leq 0.5\}} \nu & \text{ if } \phi_a \leq 0.5\\
        \max_{\nu \in \{s \in S_\epsilon: 0.5 \leq s\leq \phi_a\}} \nu & \text{ if } \phi_a \geq 0.5\\
    \end{cases}.$$
    Then, 
    \begin{align*}
        \I(\theta,\phi;H_T)
        & \overset{(a)}{=} \I(\theta,\phi,\tilde{\theta},\tilde{\phi};H_T)\\
        & \overset{(b)}{=} \I(\tilde{\phi},\tilde{\theta};H_T) + \I(\phi,\theta;H_T|\tilde{\phi},\tilde{\theta})\\
        & \overset{(c)}{\leq} \I(\tilde{\phi},\tilde{\theta};\phi, \theta) + \sum_{t=0}^{T-1}\I(\phi,\theta;A_t,O_{t+1}|\tilde{\phi},\tilde{\theta},H_t)\\
        & \overset{(d)}{\leq} \I(\tilde{\phi},\tilde{\theta};\phi, \theta) + 2T\epsilon^2 \ln \frac{1}{\epsilon^2}\\
        & \overset{(e)}{\leq} |\Ac|\ln\frac{1}{\epsilon} + 2T\epsilon^2 \ln \frac{1}{\epsilon^2}\\
        & \overset{(f)}{\leq} \left(\frac{|\Ac|}{2} + \frac{1}{2}\right)\ln 4T\\
        & \leq |\Ac|\ln(4T),
    \end{align*}
    where $(a)$ follows from the fact that $(\tilde{\theta},\tilde{\phi})\perp H_T|(\theta,\phi)$, $(b)$ follows from the chain rule of mutual information, $(c)$ follows from the data processing inequality, $(d)$ follows from Lemma \ref{le:info_env}, $(e)$ follows from the fact that $\I(\tilde{\phi},\tilde{\theta};\phi, \theta) \leq \H(\tilde{\theta},\tilde{\phi})$, and $(f)$ follows by setting $\epsilon = 1/(4T)$.
\end{proof}

\section{Bounding the Information Ratio}

Theorem \ref{th:regret_ir} establishes that to upper bound regret for an agent $\pi$, it suffices to upper bound $\E_{\pi}[\Gamma_t(\pi)]$ for that agent.  IDS minimizes $\Gamma_t$ uniformly over all histories $H_t$.  Therefore, upper bounding the regret of IDS reduces to upper bounding $\Gamma_t$ for any algorithm.  We consider the following variant of Thompson sampling to derive the regret upper bound.  For all $a \in \Ac_e$, let $n_{a,t}$ be a random variable which denotes $1 +$ the number of times action $a$ was taken up through time $t$.  For all $t$, let $\pi$ sample an action $a \in \Ac_{e}$ according to the posterior distribution $\Pr(A^*\in\cdot|H_t)$.  Then, it selects whether to query the environment or the human depending on the following conditions:
$$
    A_t =
    \begin{cases}
        a & \text{ w.p. } 1-\epsilon\\
        \bar{a} & \text{ w.p. } \epsilon
    \end{cases}.
$$
Note that $\epsilon$ is allowed to depend on $T$ since we are simply trying to upper bound the information ratio.  We now bound the regret of IDS by bounding the expected conditional information ratio of the above algorithm.  We will refer to this algorithm as $\tilde{\pi}$.

For all $a \in \Ac$, let $O_a$ denote a random variable with distribution ${\rm Bernoulli}(\phi_a)$ if $a\in\Ac_e$ and ${\rm Bernoulli}(\theta_{\bar{a}})$ if $a \in \Ac_h$.  Let $R_a = O_a\cdot O_{\bar{a}} + (1-O_a)\cdot (1-O_{\bar{a}})$ for $a \in \Ac_e$.  

\begin{lemma}
    For all $t \in \mathbb{Z}_{+}$,
    \begin{align*}
        \E_{\tilde{\pi}}\left[R^*_t - R_t|H_t\right]^2
        &\ \overset{a.s.}{\leq}\ \underbrace{2\left(\sum_{a\in \Ac_e} (1-\epsilon) \Pr(A^*=a|H_t) \cdot\left(\E\left[R_a|A^*=a,H_t\right] - \E\left[R_a|H_t\right]\right)\right)^2}_{ \Rc^{(1)}_t(\tilde{\pi})}\\
        &\quad  + \underbrace{2\left(\sum_{a\in \Ac_e} 2 \epsilon \Pr(A^*=a|H_t)\right)^2}_{\Rc^{(2)}_t(\tilde{\pi})}.
    \end{align*}
\end{lemma}
\begin{proof}
    Take all inequality and equality to hold almost surely.
    \begin{align*}
        \E_{\tilde{\pi}}\left[R^*_t - R_t|H_t\right]^2
        & = \left(\sum_{a\in \Ac} \tilde{\pi}_t(a) \cdot\left(\E\left[R_a|A^*=a,H_t\right] - \E\left[R_a|H_t\right]\right)\right)^2\\
        & = \left(\sum_{a\in \Ac_e} \Pr(A^*=a|H_t) \left((1-\epsilon)\left(\E\left[R_a|A^*=a,H_t\right] - \E\left[R_a|H_t\right]\right) + \epsilon\left(\E[R^*_t|H_t]-1\right)\right)\right)^2\\
        & \leq \left(\sum_{a\in \Ac_e} \Pr(A^*=a|H_t) \left((1-\epsilon)\left(\E\left[R_a|A^*=a,H_t\right] - \E\left[R_a|H_t\right]\right) + 2\epsilon\right)\right)^2\\
        & \overset{(a)}{\leq} 2\left(\sum_{a\in \Ac_e} (1-\epsilon) \Pr(A^*=a|H_t) \cdot\left(\E\left[R_a|A^*=a\right] - \E\left[R_a\right]\right)\right)^2 + 2\left(\sum_{a\in \Ac_e} 2 \epsilon \Pr(A^*=a|H_t)\right)^2\\
    \end{align*}
    where $(a)$ follows from the fact that $(x+y)^2 \leq 2x^2 = 2y^2$ for all $x,y\in \Re$.
\end{proof}

We now upper bound the two terms $\Rc_t^{(1)}(\tilde{\pi})$ and $\Rc_t^{(2)}(\tilde{\pi})$.

\begin{lemma}\label{le:ub1}
    For all $t \in \mathbb{Z}_+$ and $\epsilon \in [0,1]$,
    $$\Rc_{t}^{(1)}(\tilde{\pi})\ \overset{a.s.}{\leq}\ |\Ac_e|\sum_{a\in\Ac_e} (1-\epsilon)^2 \Pr(A^*=a|H_t)\cdot \I(O_a,O_{\bar{a}};\theta,\phi,A^*|H_t = H_t).$$
\end{lemma}
\begin{proof}
    Take all inequality and equality to hold almost surely.
    \begin{align*}
        &\ \Rc_{t}^{(1)}(\tilde{\pi})\\
        & \overset{(a)}{\leq} 2|\Ac|\sum_{a\in\Ac_e}(1-\epsilon)^2\Pr(A^*=a|H_t)^2\cdot\left(\E[R_a|A^*=a] - \E[R_a]\right)^2 \\
        & \overset{(b)}{\leq} 2|\Ac|\sum_{a\in\Ac_e}(1-\epsilon)^2\Pr(A^*=a|H_t)\sum_{a^*,\nu,\xi}\Pr(A^*=a^*,\theta=\nu,\phi=\xi|H_t) \left(\E[R_a|A^*=a^*,\theta=\nu,\phi=\xi] - \E[R_a]\right)^2\\
        & \overset{(c)}{\leq} |\Ac|\sum_{a\in\Ac_e}(1-\epsilon)^2\Pr(A^*=a|H_t)\cdot \I(O_a,O_{\bar{a}};\theta,\phi,A^*|H_t\leftarrow H_t)\\
        & \overset{(d)}{\leq} |\Ac|\sum_{a\in\Ac_e}(1-\epsilon)^2\Pr(A^*=a|H_t)\cdot \I(O_a,O_{\bar{a}};\theta,\phi|H_t\leftarrow H_t),
    \end{align*}
    where $(a)$ follows from Cauchy-Bunyakovsky Scharz, $(b)$ follows from the tower property and the fact that for all real-valued random variables $X$, $\E[X]^2 \leq \E[X^2]$, $(c)$ follows from Fact 9 of \cite{russo2016information}, and $(d)$ follows from the fact that $O_a\perp A^*|(\theta,\phi,H_t)$.
\end{proof}

\begin{lemma}\label{le:ub2}
    For all $t \in\mathbb{Z}_{+}$ and $\epsilon \in [0,1]$,
    $$\Rc_{t}^{(2)}(\tilde{\pi})\ \overset{a.s.}{\leq}\ 8\epsilon^2.$$
\end{lemma}
\begin{proof}
    Take all inequality and equality to hold almost surely.
    \begin{align*}
         \Rc_{t}^{(2)}(\tilde{\pi})
         & = 2\left(\sum_{a\in \Ac_e} 2 \epsilon \Pr(A^*=a|H_t)\right)^2\\
         & = 8\epsilon^2 \left(\sum_{a\in \Ac_e} \Pr(A^*=a|H_t)\right)^2\\
         & \leq 8\epsilon^2\\
    \end{align*}
\end{proof}

With these upper bounds on $\Rc^{(1)}_t(\tilde{\pi})$ and $\Rc^{(2)}_{t}(\tilde{\pi})$ we now upper bound the information ratio:

\begin{lemma}\label{le:info_ratio1}
    For all $t \in \mathbb{Z}_{+}$ and $\epsilon \in [0,1]$, 
    $$\frac{\Rc_t^{(1)}(\tilde{\pi})}{\sum_{a\in\Ac}\tilde{\pi}_t(a)\cdot\I(O_a;\theta,\phi|H_t=H_t)}\ \overset{a.s.}{\leq}\ \frac{|\Ac|(1-\epsilon)^2}{\min\{\epsilon,1-\epsilon\}}.$$
\end{lemma}
\begin{proof}
    Take all inequality to hold almost surely.
    \begin{align*}
        \frac{\Rc_t^{(1)}(\tilde{\pi})}{\sum_{a\in\Ac}\tilde{\pi}_t(a)\cdot\I(O_a;\theta,\phi|H_t=H_t)}
        &\overset{(a)}{\leq} \frac{|\Ac_e|\sum_{a\in\Ac_e}(1-\epsilon)^2\Pr(A^*=a|H_t)\cdot \I(O_a,O_{\bar{a}};\theta,\phi|H_t= H_t)}{\sum_{a\in\Ac}\tilde{\pi}_t(a) \cdot \I(O_a;\theta,\phi|H_t= H_t)}\\
        & \leq \frac{|\Ac_e|\sum_{a\in\Ac_e}(1-\epsilon)^2\Pr(A^*=a|H_t)\cdot \I(O_a,O_{\bar{a}};\theta,\phi|H_t= H_t)}{\sum_{a\in\Ac_e} \epsilon\Pr(A^*=a|H_t) \cdot \I(O_{\bar{a}};\theta,\phi|H_t= H_t)}\\
        & \overset{(b)}{\leq} \frac{2(1-\epsilon)^2}{\min\{\epsilon, 1-\epsilon\}}\frac{|\Ac_e|\sum_{a\in\Ac_e} \Pr(A^*=a|H_t) \cdot \left(\frac{1}{n_{a,t}} +\frac{1}{n_{\bar{a},t}}\right)}{\sum_{a\in\Ac_e} \Pr(A^*=a|H_t) \cdot \left(\frac{1}{n_{\bar{a},t}}+\frac{1}{n_{a,t}}\right)}\\
        & \leq \frac{|\Ac|(1-\epsilon)^2}{\min\{\epsilon,1-\epsilon\}},
    \end{align*}
    where $(a)$ follows from Lemma \ref{le:ub1} and the fact that $O_a\perp A^*|(\theta,\phi,H_t)$ and $(b)$ follows from Lemma \ref{cor:mutualinfo}.
\end{proof}

\begin{lemma}\label{le:info_ratio2}
    For all $T \in \mathbb{Z}_{+}$ and $\epsilon \in [0,1]$,
    $$\sum_{t=0}^{T-1}\frac{\Rc_t^{(2)}(\tilde{\pi})}{\sum_{a\in\Ac}\pi_t(a)\cdot\I(O_a;\theta,\phi|H_t=H_t)}\ \overset{a.s.}{\leq}\ 32\epsilon^2 T^2.$$    
\end{lemma}
\begin{proof}
    \begin{align*}
    \sum_{t=0}^{T-1}\frac{\Rc_t^{(2)}(\tilde{\pi})}{\sum_{a\in\Ac_e}\pi_t(a)\cdot\I(O_a;\theta,\phi|H_t=H_t)}
    & \overset{(a)}{\leq} \sum_{t=0}^{T-1}\frac{8\epsilon^2}{\sum_{a\in\Ac_e}\pi_t(a)\cdot\I(O_a;\theta,\phi|H_t=H_t)}\\
    & \overset{(b)}{\leq} \sum_{t=0}^{T-1}8\epsilon^2\sum_{a\in\Ac}\tilde{\pi}_t(a)\frac{1}{\I(O_{a};\theta,\phi|H_t=H_t)}\\
    & \overset{(c)}{\leq} \sum_{t=0}^{T-1}32\epsilon^2\sum_{a\in\Ac}\tilde{\pi}_t(a) n_{a,t}\\
    & \leq \sum_{t=0}^{T-1}32\epsilon^2 T\\
    & \leq 32\epsilon^2 T^2,
\end{align*}
where $(a)$ follows from Lemma \ref{le:ub2}, $(b)$ follows from Jensen's inequality, and $(c)$ follows from Corollary \ref{cor:mutualinfo}.
\end{proof}

\begin{lemma}
    For all $T \in \mathbb{Z}_{+}$ and $\epsilon \in [0,1]$,,
    $$\sum_{t=0}^{T-1}\ \E_{\pi_{\rm ids}}\left[\Gamma_t(\pi_{\rm ids})\right]\ \leq\ \frac{|\Ac|T(1-\epsilon)^2}{\min\{\epsilon,1-\epsilon\}} + 32\epsilon^2 T^2.$$
\end{lemma}
\begin{proof}
    \begin{align*}
        \sum_{t=0}^{T-1}\ \E_{\pi_{\rm ids}}\left[\Gamma_t(\pi_{\rm ids})\right]
        & = \sum_{t=0}^{T-1}\ \E_{\pi_{\rm ids}}\left[\frac{\E_{\pi_{\rm ids}}[R^*_t-R_t|H_t]}{\I_{\pi_{\rm ids}}(O_{t+1},A_t;\theta,\phi|H_t=H_t)}\right]\\
        & \overset{(a)}{\leq} \sum_{t=0}^{T-1}\ \E_{\pi_{ids}}\left[\frac{\E_{\tilde{\pi}}[R^*_t-R_t|H_t]}{\I_{\tilde{\pi}}(O_{t+1},A_t;\theta,\phi|H_t=H_t)}\right]\\
        & \overset{a.s.}{\leq} \frac{|\Ac|T(1-\epsilon)^2}{\min\{\epsilon,1-\epsilon\}} + 32\epsilon^2 T^2,
    \end{align*}
    where $(a)$ follows from the fact that IDS minimizes the information ratio uniformly for all $H_t$ and the final inequality follows from Lemmas \ref{le:info_ratio1} and \ref{le:info_ratio2}.
\end{proof}

\section{IDS Regret Bound}

With bounds on the information ratio and the mutual information $\I(\theta,\phi;H_T)$, we derive the following regret bounds:

\begin{theorem}\label{th:regret_bound1}
    For all $T\in\mathbb{Z}_{+}$ and $\epsilon \geq 0$,
    $$\Re(\pi_{\rm ids}, T)\ \leq\ \sqrt{\frac{|\Ac|T(1-\epsilon)^2}{\min\{\epsilon,1-\epsilon\}} + 32\epsilon^2 T^2} \cdot \sqrt{|\Ac|\ln 4T}.$$
\end{theorem}

\idsRegret*
\begin{proof}
    First consider the case in which $T \leq |\Ac|$.  Then we have that
    \begin{align*}
        \Re(\pi_{\rm ids}, T)
        & \overset{(a)}{\leq} \sqrt{\frac{|\Ac|T(1-\epsilon)^2}{\min\{\epsilon,1-\epsilon\}} + 32\epsilon^2 T^2} \cdot \sqrt{|\Ac|\ln 4T}\\
        & \leq \sqrt{\frac{|\Ac|T(1-\epsilon)^2}{\min\{\epsilon,1-\epsilon\}} + 32\epsilon^2 |\Ac|^{\frac{1}{2}}T^{\frac{3}{2}}} \cdot \sqrt{|\Ac|\ln 4T}\\
        & \overset{(b)}{\leq} \sqrt{\frac{|\Ac|T(|\Ac|^{-\frac{1}{2}})^2}{\min\{1-|\Ac|^{-\frac{1}{2}},|\Ac|^{-\frac{1}{2}}\}} + 32 |\Ac|^{\frac{1}{2}}T^{\frac{3}{2}}} \cdot \sqrt{|\Ac|\ln 4T}\\
        & \leq \sqrt{|\Ac|T(|\Ac|^{-\frac{1}{2}}) + 32 |\Ac|^{\frac{1}{2}}T^{\frac{3}{2}}} \cdot \sqrt{|\Ac|\ln 4T}\\
        & = \sqrt{(|\Ac|^{3/2}T+32|\Ac|^{3/2}T^{\frac{3}{2}})\ln(4T)}\\
        & \leq \sqrt{33\ln (4T)}|\Ac|^{\frac{3}{4}}T^{\frac{3}{4}},
    \end{align*}
    where $(a)$ follows from Theorem \ref{th:regret_bound1} and $(b)$ follows by setting $1-\epsilon = |\Ac|^{-1/2}$.

    Now consider the case in which $T > |\Ac|$.  Then we have that
    \begin{align*}
        \Re(\pi_{\rm ids}, T)
        & \overset{(a)}{\leq} \sqrt{\frac{|\Ac|T(1-\epsilon)^2}{\min\{\epsilon,1-\epsilon\}} + 32\epsilon^2 T^2} \cdot \sqrt{|\Ac|\ln 4T}\\
        & \leq \sqrt{\frac{|\Ac|T(1-\epsilon)}{\min\{\epsilon,1-\epsilon\}} + 32\epsilon T^2} \cdot \sqrt{|\Ac|\ln 4T}\\
        & \leq \sqrt{\frac{|\Ac|T\left(1-\frac{|\Ac|^{1/2}}{T^{1/2}}\right)}{\min\left\{\frac{|\Ac|^{1/2}}{T^{1/2}},1-\frac{|\Ac|^{1/2}}{T^{1/2}}\right\}} + 32\frac{|\Ac|^{1/2}}{T^{1/2}} T^2} \cdot \sqrt{|\Ac|\ln 4T}\\
        & \leq \sqrt{|\Ac|^{\frac{1}{2}}T^{\frac{3}{2}} + 32|\Ac|^{\frac{1}{2}}T^{\frac{3}{2}}}\cdot\sqrt{|\Ac|\ln(4T)}\\
        & = \sqrt{33\ln (4T)}|\Ac|^{\frac{3}{4}}T^{\frac{3}{4}},
    \end{align*}
    where $(a)$ follows from Theorem \ref{th:regret_bound1} and $(b)$ follow by setting $\epsilon = \frac{|\Ac|^{1/2}}{T^{1/2}}$.
\end{proof}

\end{document}